\def\year{2020}\relax
\documentclass[letterpaper]{article} 
\usepackage{aaai}  
\usepackage{times}  
\usepackage{helvet} 
\usepackage{courier}  
\usepackage[hyphens]{url}  
\usepackage{graphicx} 
\usepackage{graphicx}  
\usepackage{amsmath}
\usepackage{amssymb}
\title{PuckNet: Estimating hockey puck location from broadcast video}

\author{Kanav Vats \quad William McNally \quad Chris Dulhanty \quad  Zhong Qiu Lin \\ \Large{\textbf{David A. Clausi \quad John Zelek}}\\
University of Waterloo \hspace{2cm}\\
Waterloo, Ontario, Canada\\
{\tt\small \{k2vats,wmcnally,chris.dulhanty,zhong.q.lin,dclausi,jzelek\}@uwaterloo.ca}}
 
\begin{document}

\maketitle

\begin{abstract}
\begin{quote}
Puck location in ice hockey is essential for hockey analysts for determining the location of play and analyzing game events. However, because of the difficulty involved in obtaining accurate annotations due to the extremely low visibility  and commonly occurring occlusions of the puck, the problem is very challenging. The problem becomes even more challenging in broadcast videos with changing camera angles. We introduce a novel methodology for determining puck location from approximate puck location annotations in broadcast video. Our method uniquely leverages the existing puck location information that is publicly available in existing hockey event data and uses the corresponding one-second broadcast video clips as input to the network. The rationale behind using video as input instead of static images is that with video, the temporal information can be utilized to handle puck occlusions. 
The network outputs a heatmap representing the probability of the puck location using a 3D CNN based architecture. The network is able to regress the puck location from broadcast hockey video clips with varying camera angles. Experimental results demonstrate the capability of the method, achieving 47.07\% AUC on the test dataset. The network is also able to estimate the puck location in defensive/offensive zones with an accuracy of greater than 80\%.
\end{quote}
\end{abstract}

\section{Introduction}
Ice hockey is played by an estimated 1.8 million people worldwide~\cite{iihf2018survey}. As a team sport, the positioning of the players and puck on the ice are critical to offensive and defensive strategy~\cite{thomas2006impact}. Currently, practical methods for tracking the position of each player and the puck for the full duration of a hockey match are limited. Advances in computer vision have shown promise in this regard~\cite{lu2009tracking,pidaparthy2019keep}, but ultimately remain in the developmental phase. As an alternative, radio-frequency identification is currently being explored for player and puck tracking~\cite{cavallaro1997foxtrax,nhl2019tracking}, but may only be financially and logistically feasible at the most elite professional level, e.g., the National Hockey League (NHL). Information regarding player and puck position is therefore inaccessible in most cases. As a result, the conventional heuristic approach for evaluating the effectiveness of team strategies involves analyzing the record of \textit{events} that occurred during the match (turnover, shot, hit, face-off, dump, etc.)~\cite{tora2017classification,fani2017hockey}.

\begin{figure}[t]
\begin{center}
\includegraphics[width=\linewidth]{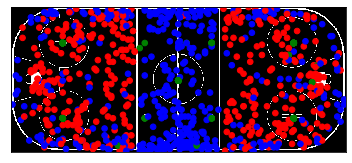}
\end{center}
   \caption{Distribution of some puck locations on the hockey rink. The locations are evenly distributed throughout the ice rink. The red, blue and green circles correspond to the puck locations of shots, dumps and faceoffs  respectively.}
\label{fig:event_distibution}
\end{figure}

In the NHL, events are recorded on a play-by-play basis by dedicated statisticians\footnote{Play-by-play event data is publicly available for all NHL games at \url{NHL.com}}. Additionally, third-party hockey analytics companies 
provide more in-depth event information, including a greater number of event types and event details, for the NHL and other hockey leagues around the world. Each event is linked with a game-clock timestamp (1-second resolution), and an approximate location where the event occurred on the rink. Generally speaking, the event location corresponds to the approximate location of the puck. Therefore, there exists an expansive knowledgebase of approximate puck location information that has, until now, not been exploited. To this end, this paper explores the following idea: \textit{can we leverage existing hockey event annotations and corresponding broadcast video to predict the location of the puck on the ice?}

\begin{figure}[t]
\begin{center}
\includegraphics[width=\linewidth]{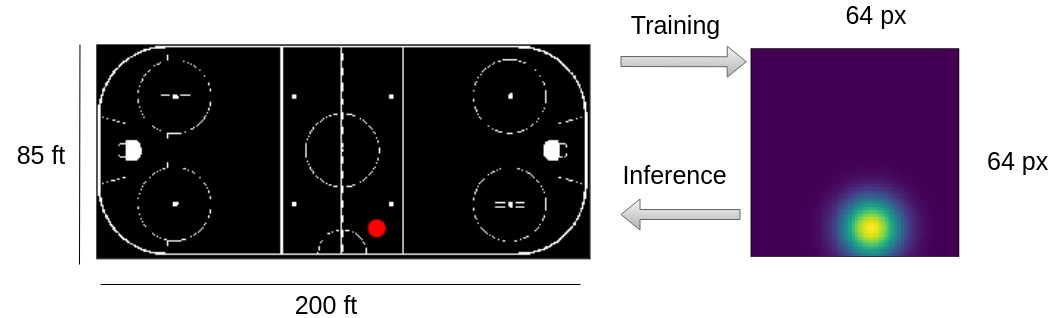}
\end{center}

   \caption{Illustration of the scaling transformation used to transform the puck annotation from the ice rink coordinates to the heatmap coordinates. For training, the annotations are transformed from the ice rink coordinates to the heatmap coordinates, whereas, predicated heatmap is transformed to ice-rink coordinates for inference.}
\label{figure:transformation}
\end{figure}

Using a relatively small dataset of hockey events containing approximate puck locations (distribution shown in Figure ~\ref{fig:event_distibution}), we use a 3D CNN to predict the puck position in the rink coordinates using the corresponding 1-second broadcast video clips as input. As such, the 3D CNN is tasked with simultaneously (1) localizing the puck in RGB video and (2) learning the homography between the broadcast camera and the static rink coordinate system. To our best knowledge, this represents a novel computer vision task that shares few similarities with any existing tasks. Drawing inspiration from the domain of human pose estimation, we model the approximate spatial puck location using a 2D Gaussian, as shown in Figure \ref{figure:transformation}. 

\section{Background}
Pidaparthy and Elder \shortcite{pidaparthy2019keep} proposed using a CNN to regress the puck's pixel coordinates from single high-resolution frames collected via a static camera for the purpose of automated hockey videography. Estimating the puck location from a single frame is a challenging task due to the relatively small size of the puck compared to the frame, occlusions from hockey sticks, players, and boards, and the significant motion blur caused by high puck velocities. Furthermore, their method was not based on existing data and thus required extensive data collection and manual annotation.

Remarking that humans can locate the puck position from video with the help of contextual cues and temporal information, our method incorporates temporal information in the form of RGB video to help localize the puck. Additionally, our method differs from Pidaparthy and Elder in that we use puck location information obtained from existing hockey event data, and directly learn the camera-rink homography instead of using a manual calibration.

\begin{figure*}[t]
\begin{center}
\includegraphics[width=.8\linewidth, height =.25\linewidth]{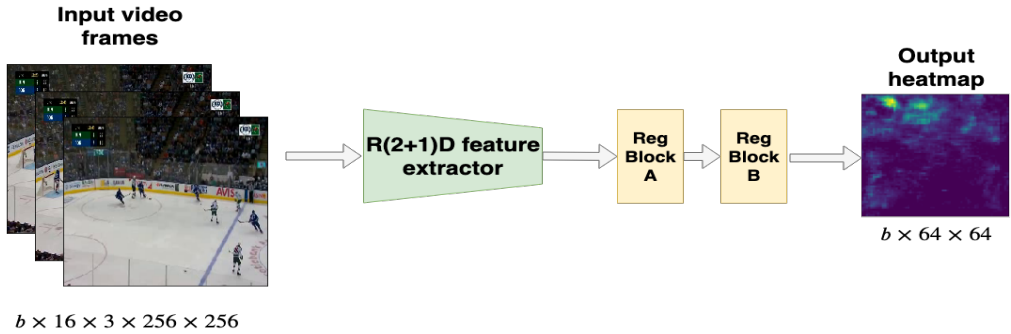}
\end{center}
   \caption{The overall network architecture. Input tensor of dimension $b\times16\times3\times256\times256$ ($b$ denotes the batch size) is input into the R(2+1)D feature extractor consisting of the first nine layers of the R(2+1)D network. The feature extractor outputs $b\times8\times128\times64\times64$ tensor representing the intermediate features. The intermediate features are finally input into two regression blocks. The first regression block(Reg Block A) outputs a $b\times2\times32\times64\times64$ tensor while the second regression block outputs the final predicted heatmap. }
\label{figure:overall_net}
\end{figure*}

\section{Methodology}
\subsection{Dataset}
The dataset consists of 2716, 60 fps broadcast NHL clips with an original resolution of $1280 \times 720$ pixels of one second each with the approximate puck location annotated. The videos are resized to a dimension of $256 \times 256$ pixels for computation. The puck locations are evenly distributed throughout the ice rink as can be seen from Figure \ref{fig:event_distibution}. The dataset is split such that 80\% of the data is used for training, 10\% for validation and 10\% for testing. 

\subsection{Experiment}

\begin{figure}[t]
\begin{center}
\includegraphics[width=\linewidth, height =.6\linewidth]{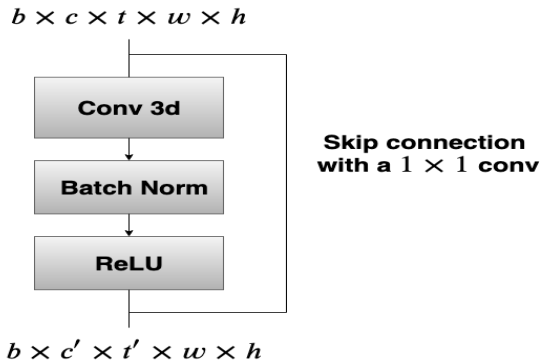}
\end{center}
   \caption{Illustration of the regression block applied after the R(2+1)D network backbone. The input and outputs are 5D tensors where $b,c,t,w$ and $h$ denote batch size, number of channels, temporal dimension, width and height of the feature map respectively. Here $c^{'}<x$ and $t^{'}<t$ since the number of channels and timesteps have to be reduced so that a single heatmap can be generated.}
\label{fig:regression_block}
\end{figure}

We use the 18 layer R(2+1)D \cite{tran} network pretrained on the Kinetics dataset  \cite{Kay2017TheKH} as a backbone for regressing the puck location from video.  The input to the network consists of $16$ video frames $\{I_{i} \in R^{3 \times 256 \times 256}\:|\: i \in [1,..,16]\}$ sampled from a one second video clip. The $16$ frames are sampled from a uniform distribution. For preprocessing, the image frame RBG pixel values are scaled to the $[0,1]$ range and normalized by the Kinetics dataset mean and standard deviation. The features maps obtained from the 9th layer of the R(2+1)D network is fed into two RegressionBlocks illustrated in Figure \ref{fig:regression_block}.  The first five layers of the R(2+1)D network are kept frozen during training in order to reduce the computational cost and maintain a batch size of 10 on a single GPU machine. Each regression block consists of a 3D convolutional layer, batch normalization and ReLU non-linearity. The final output of the network is a two-dimensional heatmap $h \in R^{64 \times 64}$ representing the probability distribution of the puck location. We chose a heatmap based approach instead of directly regressing the puck coordinates in order to account for the uncertainty in the ground truth annotations. The overall network architecture is illustrated in Figure \ref{figure:overall_net} and Table \ref{table:network}. The ground truth heatmap consists of a Gaussian with mean $\mu$ equal to the ground truth puck location and standard deviation $\sigma$. Mean squared error (MSE) loss between the ground truth and predicted heatmap is minimized during training. \par
\begin{figure}[t]
\begin{center}
\includegraphics[width=.5\linewidth, height =.5\linewidth]{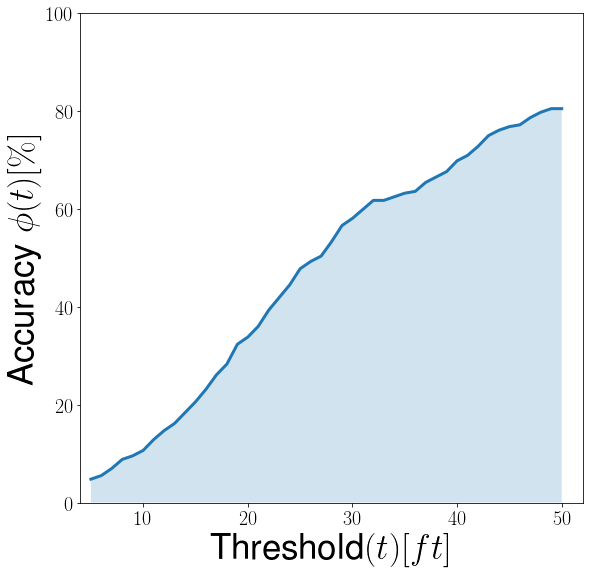}
\end{center}
   \caption{Overall AUC for the best performing model with random sampling and $\sigma = 25$.}
\label{figure:AUC}
\end{figure}

The size of the NHL hockey rink is $200ft \times 85ft$. In order to predict a $64\times64$ dimensional square heatmap, a scaling transformation $\tau: R^{200\times85} \rightarrow R^{64\times64}$  is applied to the ground truth puck annotations in rink coordinates while training. Let $hmap\_width$ and $hmap\_height$ denote the output heatmap width and height respectively. The transformation matrix is given by: $$ \tau =
\begin{pmatrix} 
\frac{hmap\_width}{200} & 0 & 0 \\
0 & \frac{hmap\_height}{85} & 0 \\
0 & 0 & 1 \\
\end{pmatrix}$$ During testing, inverse transformation $\tau^{-1}$ is applied to convert back to the rink coordinates. This process is illustrated in Figure \ref{figure:transformation}. \par
We use the Adam optimizer with an initial learning rate of .0001 with a batch size of 10. We use the Pytorch 1.3 framework on an Nvidia GTX 1080Ti GPU.

\section{Results and Discussion}
\begin{figure}[t]
\begin{center}
\includegraphics[width=.5\linewidth, height =.5\linewidth]{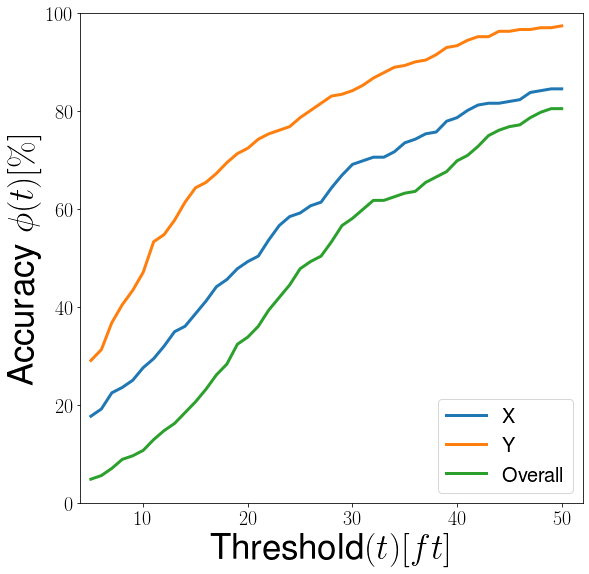}
\end{center}
   \caption{The accuracy curves corresponding to the best performing model.}
\label{figure:AUC1}
\end{figure}

\begin{table}[!t]
    \centering 
    \caption{Network architecture. k,s and p denote kernel dimension, stride and padding respectively. $Ch_{i}$ and $Ch_{o}$ and $b$ denote the number of channels going into and out of a block and batch size respectively. }
    \footnotesize
    \setlength{\tabcolsep}{0.2cm}
    \begin{tabular}{|c|}
    \hline \textbf{Input} $b\times16\times3\times256\times256$  \\\hline\hline
     \textbf{Feature extractor}\\
     First $9$ layers of R(2+1)D network
 \\ \hline
 \textbf{RegBlock A} \\
 Conv3D \\
 $Ch_{i} = 128, Ch_{o} = 32$ \\
   (k = $4\times1\times1$,
      s = $4\times1\times1$,
   p = 0) \\
      Batch Norm 3D    \\
     ReLU  \\ \hline
     \textbf{RegBlock B} \\
  Conv3D \\
   $Ch_{i} = 32, Ch_{o} = 1$ \\
   (k = $2\times1\times1$,
      s = $2\times1\times1$,
   p = 0) \\
      Batch Norm 3D    \\
      ReLU
    \\ \hline
        \textbf{Output} $b\times64\times64$ \\ \hline
    \end{tabular}
    \label{table:network}
\end{table}

\subsection{Accuracy Metric}
A test example is considered to be correctly predicted at a tolerance $t$ feet if the L2 distance between the ground truth puck location $z$ and predicted puck location $z_{0}$ is less than $t$ feet. That is $||z - z_{0}||_{2}<t$. Let $\phi(t)$ denote the percentage of examples in the test set with correctly predicted position puck position at a tolerance of $t$. We define the accuracy metric as the area under the curve (AUC) $\phi(t)$   at tolerance of $t=5$ feet to $t=50$ feet.

\begin{figure}[t]
\begin{center}
\includegraphics[width=.6\linewidth, height =.3\linewidth]{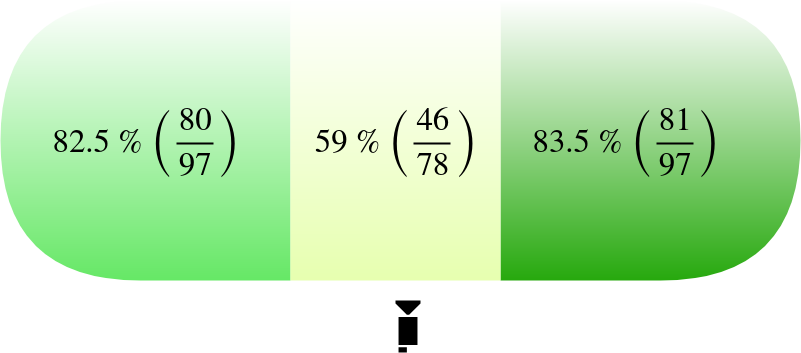}
\end{center}
   \caption{Zone-wise accuracy. The figure represents the hockey rink with the text in each zone represents the percentage of test examples predicted correctly in that zone. The position of the camera is at the bottom.  }
\label{figure:zone_3}
\end{figure}

\begin{figure}[t]
\begin{center}
\includegraphics[width=.6\linewidth, height =.3\linewidth]{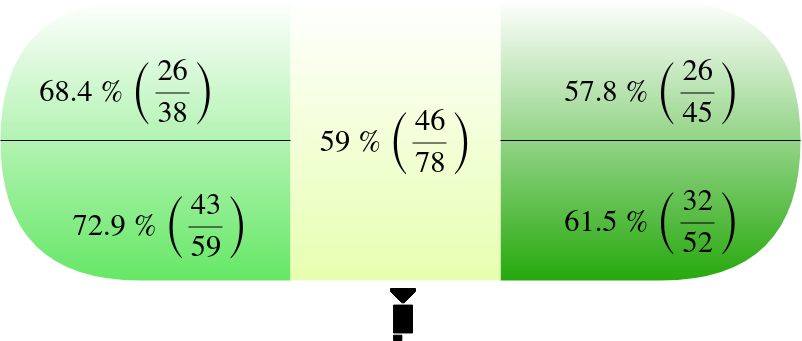}
\end{center}
   \caption{Zone-wise accuracy with offensive and defensive zones further split into two. The figure represents the hockey rink with the text in each zone represents the percentage of test examples predicted correctly in that zone. The position of the camera is at the bottom.}
\label{figure:zone_3_split}
\end{figure}

\subsection{Discussion}
Figure \ref{figure:AUC} shows the variation of overall accuracy with tolerance $t$ for the best performing model trained with $\sigma = 25$. The accuracy increases almost linearly reaching $\sim60\%$ accuracy for $t=30$ feet. The AUC score for the model is $47.07 $~\%. Figure \ref{figure:AUC1} shows the accuracy vs tolerance plot for the $\sigma =25$ model, in the horizontal(X) and vertical(Y) directions separately. The model is able to locate the puck position with the highest accuracy in Y(vertical) direction reaching an accuracy of $\sim65\%$ at a tolerance of $t=15$ feet. This is because the vertical axis is more or less always visible in the camera field of view. This cannot be said for the horizontal(X) direction since the camera pans horizontally and hence, the models has to learn the viewpoint changes. \par

\begin{table}[!t]
    \centering
    \caption{AUC values for different values of $\sigma$.}
    \footnotesize
    \setlength{\tabcolsep}{0.15cm}
    \begin{tabular}{c|c|c|c}\hline
      $\sigma$ & AUC(overall) & AUC(X)  & AUC(Y)   \\\hline\hline
        10 & 36.85 & 48.84 & 72.25 \\ 
       15 & 42.51 & 53.86 &  \textbf{77.12}\\
         20 & 45.80 & 57.31 & 76.66 \\ 
        25 & \textbf{47.07} & \textbf{58.85} & 76.78 \\
        30 & 42.86 & 54.23 & 76.76 \\
    \end{tabular}
    \label{table:results}
\end{table}

Table \ref{table:results} shows the variation of AUC for different  values of $\sigma$. The highest AUC score achieved is corresponding to $\sigma = 25$ ($47.07 $~\%). A lower value of $\sigma$ results in a lower accuracy. A reason for this can be that with lower $\sigma$, the ground truth Gaussian distribution becomes more rigid/peaked, which makes learning difficult. For a value of $\sigma>25$, the accuracy again lowers because the ground truth Gaussian becomes very spread out, which lowers accuracy on lower tolerance levels. \par
Two kinds of sampling techniques were investigated: 1) Random sampling from a uniform distribution 2) Constant interval sampling at an interval of 4 frames. Random sampling outperforms uniform sampling because it acts as a form of data augmentation. This is shown in Table \ref{table:sampling}. \par

Figure \ref{figure:zone_3} shows the zone-wise accuracy of the model. A prediction is labelled as correct if it lies in the same zone as the ground truth. The model shows good performance in the offensive and defensive zones with an accuracy greater than $80\%$. The model maintains reasonable performance when the defensive and offensive zones are further split into two (Figure \ref{figure:zone_3_split}).

\section{Conclusion and Future Work}
We have presented a novel method to locate the approximate puck position from video. The model can be used to know the zone in which the puck was present at a particular moment in time, which can be of practical significance to know the exact location of play and as a prior information for recognizing game events. The results obtained are preliminary and in the future more cues such as player detections, player trajectories on ice and optical flow can be taken into account to obtain more accurate results. It would also be interesting to apply the proposed methodology in sports such as soccer.

\begin{table}[!t]
    \centering
    \caption{Comparison between uniform and random sampling settings. Random sampling outperforms uniform sampling because it acts as a form of data augmentation.}
    \footnotesize
    \setlength{\tabcolsep}{0.15cm}
    \begin{tabular}{c|c|c|c|c}\hline
  
      Sampling  & $\sigma$ & AUC(overall) & AUC(X)  & AUC(Y)   \\\hline\hline
        Random & 20 & 45.80 & 57.31 & 76.66 \\ 
       Constant interval & 20 & 36.55 & 49.24 & 71.41 \\

    \end{tabular}
    \label{table:sampling}
\end{table}
\section{Acknowledgment}
This work was supported by Stathletes through the Mitacs Accelerate Program and Natural Sciences
and Engineering Research Council of Canada (NSERC).

\bibliographystyle{aaai}
\bibliography{aaai}

\end{document}